\documentclass[9pt,sigconf]{acmart}

\usepackage{multirow}
\usepackage[utf8]{inputenc}
\usepackage{cleveref}
\crefname{section}{§}{§§}
\Crefname{section}{§}{§§}
\usepackage{float}
\usepackage{graphicx}
\usepackage{hyperref}
\usepackage{color}
\hypersetup{
	colorlinks=true,
	linkcolor=red,
	citecolor=blue,
	filecolor=black,
	urlcolor=blue,
}

\newcommand{\tool}{{\tt apk2vec}}

\newcommand{\soa}{state-of-the-art}
\newcommand{\wv}{{\tt word2vec}}
\newcommand{\dv}{{\tt doc2vec}}

\newcommand{\sv}{{\tt sub2vec}}
\newcommand{\svn}{{\tt sub2vec\_N}}
\newcommand{\svs}{{\tt sub2vec\_S}}
\newcommand{\gv}{{\tt graph2vec}}
\newcommand{\sss}{{substructure}}
\newcommand{\sg}{subgraph}
\newcommand{\apk}{{\textit{apk}}}

\newcommand{\wlk}{{\tt WLK}}
\newcommand{\gefsg}{{\tt GE-FSG}}

\usepackage[ruled,vlined,linesnumbered]{algorithm2e}

\newtoks\therules
\therules={}
\def\appendto#1#2{\expandafter#1\expandafter{\the#1#2}}
\def\gobblefirst#1{
	#1\expandafter\expandafter\expandafter{\expandafter\@gobble\the#1}}%
\def\LState{\State\unskip\the\therules}
\def\printindent{\unskip\the\therules}%

\usepackage{hyperref}
\usepackage{color}
\hypersetup{
	colorlinks=true,
	linkcolor=red,
	citecolor=blue,
	filecolor=black,
	urlcolor=blue,
}
\usepackage{enumerate}
\usepackage{enumitem}
\usepackage{footmisc}
\usepackage{footnote}
\usepackage[english]{babel}
\usepackage{blindtext}

\usepackage{etoolbox}

\usepackage{amsmath,amssymb,amsfonts}
\usepackage{algorithmic}

\begin{document}
\sloppy

\title{\tool: Semi-supervised multi-view representation learning for profiling Android applications}
\author{Annamalai Narayanan*, Charlie Soh*, Lihui Chen, Yang Liu and Lipo Wang}
\email{[annamala002, csoh004]@e.ntu.edu.sg, [elhchen, yangliu,elpwang]@ntu.edu.sg}
\affiliation{%
	\institution{Nanyang Technological University, Singapore}
}
%

\begin{abstract}
	Building behavior profiles of Android applications (apps) with holistic, rich and multi-view information (e.g., incorporating several semantic views of an app such as API sequences, system calls, etc.) would help catering downstream analytics tasks such as app categorization, recommendation and malware analysis  significantly better. Towards this goal, we design a semi-supervised Representation Learning (RL) framework named \tool{} to automatically generate a compact representation (\textit{aka} profile/embedding) for a given app. More specifically, \tool{} has the three following unique characteristics which make it an excellent choice for large-scale app profiling: (1) it encompasses information from multiple semantic views such as API sequences, permissions, etc., (2) being a semi-supervised embedding technique, it can make use of labels associated with apps (e.g., malware family or app category labels) to build high quality app profiles, and (3) it combines RL and feature hashing which allows it to efficiently build profiles of apps that stream over time (i.e., online learning).
	
	The resulting semi-supervised multi-view hash embeddings of apps could then be used for a wide variety of downstream tasks such as the ones mentioned above. 
	{\color{black}Our extensive evaluations with more than 42,000 apps demonstrate that \tool's app profiles could significantly outperform \soa{} techniques in four app analytics tasks namely, malware detection, familial clustering, app clone detection and app recommendation.  
	}
\end{abstract}


\maketitle

\section{Introduction}
\label{sec:intro}


The low threshold for entering the official and third-party Android app markets has attracted a large number of developers, resulting in an exponential growth of apps. At the point of this writing, Google play \cite{gp} is hosting more than 3.6 million apps \cite{appbrain}, with thousands of new apps being added daily. Due to the wide range of functionalities provided by the apps (e.g., online shopping, banking, etc.), they have become an indispensable part of people's daily life. However, this astronomical volume of functionality rich apps, has raised several challenging issues. A few significant ones are as follows: (i) app markets are facing difficulties in organizing large volumes of diverse apps to allow convenient and systematic browsing by the users, (ii) due to the rapid growth rate in app volumes, it is becoming increasingly tough for markets to recommend up-to-date and meaningful  apps that matches users' search queries, and (iii) with a significant number of plagiarists and malware authors hidden among app developers, these markets have been plagued with app clones and malicious apps. 

One could observe that a systematic and deep understanding of apps'
behaviors is essential to solve the aforementioned issues. Building high-quality behavior profiles of apps could help in determining the semantic similarity among the apps, which is pivotal to addressing these issues. Recent research \cite{kaiclones,casandra,mkldroid,dapasa,faldroid,adagio,androidvuln,droidsift,tiantdse,massvet} reveals that compared to primitive representations of programs (e.g., counts of system-calls,  Application Programming Interfaces (APIs) used etc.) graph representations (e.g., Control Flow Graphs (CFGs), call graphs, etc.) are ideally suited for app profiling, as the latter retain program semantics well, even when the apps are obfuscated.  
Reinforcing this fact, many recent works achieved excellent results using graph representations along with Machine Learning (ML) techniques on a plethora of program analytics tasks such as malware detection \cite{droidsift,adagio,casandra,mkldroid,massvet}, familial classification \cite{faldroid}, clone detection \cite{kaiclones,wlkclones}, library detection \cite{phalib} etc. In effect, these works cast their respective program analytics task as a graph analytics task and apply existing graph mining techniques \cite{adagio} to solve them. 
Typically, these ML algorithms work on vectorial representations (\textit{aka} embeddings) of graphs. Hence, arguably, one of the most important factors that determines the efficacy of these downstream analytics tasks is the quality of such embeddings.

Besides the choice of graph representations, another pivotal factor that influences the aforementioned tasks are the features that could be extracted from them. In the case of app analytics, the most prominent features in recent literature include API/system-call sequences observed \cite{casandra}, permissions \cite{drebin} and information source/sinks used \cite{susi}, etc.  Evidently, each of these feature sets provides a different semantic \textit{perspective} (interchangeably referred as \textit{view}) of the apps' behavior with different inherent strengths and limitations. As revealed by existing works \cite{drebin,mkldroid}, capturing multiple semantic views with different modalities would help to improve the accuracy of downstream tasks significantly. Furthermore, any form of labeling information (e.g., malware family label, app category label, etc.) could be of significant help in building semantically richer and more accurate app profiles. 

Towards catering the above-mentioned applications, in this work, we propose a Representation Learning (RL) technique to build data-driven, compact and versatile behavior profiles of apps. Based on the above observations, the following challenges have to be addressed to obtain such a profile:\\
\textbf{(C1) Handcrafted features.} Graph representations of programs such as CFGs are highly expressive data structures. Consequently, representing them as vectors without losing much of their expressiveness is non-trivial. A typical solution is to use graph kernels \cite{graphlet,wlk,rw,sp} which leverage on graph \sss s (e.g., shortest paths, graphlets etc.) to build graph embeddings. However, these substructures are \textit{handcrafted}. Therefore, when used on large datasets, these features lead to building high dimensional, sparse and non-smooth graph embeddings which do not generalize well and thus yield suboptimal accuracies \cite{dgk,sg2vec}. 
\\
\textbf{(C2) Fully supervised or unsupervised RL.} Addressing the limitations of graph kernels, several data-driven supervised (e.g., CNNs \cite{patchy}, RNNs \cite{gam}) and unsupervised (e.g.,skipgrams \cite{sub2vec,g2v}, autoencoders \cite{skipgraph}) graph embedding approaches have been proposed. Both types of approaches exhibit good generalization and offer excellent accuracies. However, the supervised embedding methods suffer from the following limitations: (i) they require typically large volumes of labeled graphs to learn meaningful embeddings, which is undesirable and often impractical for large-scale app analytic tasks, and (ii) the embeddings thus learnt are specific to one particular analytics task and may not be transferable to others.
On the other hand, unsupervised embedding methods do not exhibit these limitations. However, in many cases, a portion of the dataset may have labels or some samples may have more than one label (e.g., an app may have several labels such as category to which it belongs, whether or not it is malicious, etc.). Unsupervised embedding approaches are incapable of leveraging such labels which contain valuable semantic information.
\\
\textbf{(C3) Scalability.} Though RL based methods provide graph embeddings which generalize well, when used on large datasets, they exhibit poor scalability both in terms of memory and time requirements. This is because they have extremely large number of parameters to train (especially in models such as RNNs and skipgrams). 
\\
\textbf{(C4) Integrating information from multiple views.} All the above mentioned approaches are typically designed to capture only one semantic view of the program through their embeddings. This severely limits their potentials to cater to a wide range of downstream tasks. Effectively integrating information from multiple views is challenging, but of paramount importance in building comprehensive embeddings capable of catering to a variety of tasks. 


\textbf{Our approach.}
Driven by these motivations, we develop a static analysis based semi-supervised multi-view RL framework named \tool{} to build high-quality data-driven profiles of Android apps. \tool{} has two major phases: (1) \textit{a static analysis phase} in which a given \apk{} file is disassembled and three different dependency graphs (DGs), each representing a distinct semantic view are extracted and, (2) \textit{an embedding phase} in which a neural network is used to combine the information from these three DGs and label information (if available) to learn one succinct embedding for the \textit{apk}. To this end, \tool{} combined and extends several \soa{} RL ideas such as multimodal (\textit{aka}  multi-view) RL, semi-supervised neural embedding and feature hashing.

\tool{} addresses the above-mentioned challenges in the following ways:
\begin{itemize}
	[leftmargin=*]
	\setlength\itemsep{0em}
	
	\item \textbf{Data-driven embedding:} Unlike graph kernels, \tool{} uses a skipgram neural network \cite{w2v,g2v} that automatically learns features from large corpus of graph data to produce high quality dense embeddings. This in effect addresses challenge C1. 
	
	\item \textbf{Semi-supervised task-agnostic embedding:}  \tool's neural network facilitates using class labels of \apk{s} (incl. multiple labels per sample) if they are available to build better app profiles. However, these embeddings are still task-agnostics and hence can be used for a variety of downstream tasks. This helps addressing challenge C2.
	
	\item \textbf{Hash embedding:} Recently, Svenstrup et al \cite{he} proposed a scalable feature hashing based word embedding model which required much lesser number of trainable parameters than conventional RL models. Besides this improvement in efficiency, hash embedding model also facilitates learning embeddings when instances stream over time. Inspired by this idea, in \tool, we develop an efficient hash embedding model for graph/subgraph embedding, thus addressing challenge C3.
	
	\item \textbf{Multi-view embedding:} \tool's neural network facilitates multimodal RL through a novel learning strategy (see \S \ref{ss:ep}). This helps to integrate three different DGs that emerge from a given \apk{} file and produce one common embedding. Thus \tool{} facilitates combining information from different views in a systematic and non-linear manner, thereby addressing challenge C4.

\end{itemize}


\textbf{Experiments.} To evaluate our approach, we perform a series of experiments on various app analytics tasks (incl. supervised, semi-supervised and unsupervised learning tasks), using a dataset of more than 42,000 real-world Android apps. 
The results show that our semi-supervised multimodal embeddings can achieve significant improvements in terms of accuracies over unsupervised/unimodal RL approaches and graph kernel methods while maintaining comparable efficiency. The improvements in prediction accuracies range from 1.74\% to 5.93\% (see \S {\ref{sec:eval}} for details).

In summary, we make the following contributions:

\begin{itemize}
	[leftmargin=*]
	\setlength\itemsep{0em}
	\item We propose \tool{}, a static analysis based data-driven semi-supervised multi-view graph embedding framework, to build task-agnostic profiles for Android apps (\S \ref{sec:meth}). To the best of our knowledge, this is the first app profiling framework that has three aforementioned unique characteristics.
	
	\item We propose a novel variant of the skipgram model by introducing a view-specific negative sampling technique which facilitates integrating information from different views in a non-linear manner to obtain multi-view embeddings (\S \ref{ss:ns}).
	
	\item We extend the feature hashing based word embedding model to learn multi-view graph/subgraph embeddings. Hash embeddings improve \tool's overall efficiency and support online RL  (\S \ref{ss:he}).
	
	\item We make an efficient implementation of \tool{} and the profiles of all the apps used in this work publicly available at \cite{ourweb}.
	
	
\end{itemize}


\section{Problem Statement}

Given a set of \apk{s} $\mathbb{A} = \{a_1, a_2, ...\}$, a set of corresponding labels $\mathbb{L} = \{l_1,l_2,...\}$ (some of which may be empty i.e., $\forall l_i \in \mathbb{L}, |l_i| \ge 0$) for each app in $\mathbb{A}$ and a positive integer $\delta$ (i.e., embedding size), we intend to learn $\delta$-dimensional distributed representations for every \apk{ }$a_i \in \mathbb{A}$. The matrix representations of all \apk{s} is denoted as $\Phi^\mathbb{A} \in \mathbb{R}^{|{\mathbb{A}|} \times \delta}$.

More specifically, $a_i \in \mathbb{A}$ can be represented as a three-tuple: $a_i = (G_i^v)$ where $v \in \{A, P, S\}$ and $G_i^A, G_i^P, G_i^S$ denote its API Dependency Graph (ADG), Permission Dependency Graph (PDG), and information Source \& sink Dependency Graph (SDG), respectively (refer to \S \ref{ss:sa} for details on constructing these DGs). Furthermore, a DG can be represented as $G_i^v = (N_i^v, E_i^v, \lambda^v)$, where $N_i^v$ is the set of nodes and $E_i^v \subseteq N_i^v \times N_i^v $ is the set of edges in $G_i^v$. A labeling function $\lambda^v:N_i^v \to L^v$ assigns a label to every node in $N_i^v$ from  alphabet set $L^v$. 

Given $G^v = (N^v, E^v, \lambda^v)$ and $sg^v = (N_{sg}^v, E_{sg}^v, \lambda_{sg}^v)$. $sg^v$ is a subgraph of $G$ iff there exists an injective mapping $\mu:N_{sg}^v \to N^v$ such that $(n_1,n_2) \in E_{sg}^v$ iff $(\mu(n_1), \mu(n_2)) \in E^v$. In this work, by subgraph, we strictly refer to a specific class of subgraphs, namely, rooted subgraphs. In a given graph $G^v$, a rooted subgraph of degree $d$ around node $n \in N^v$ encompasses all the nodes (and corresponding edges) that are reachable in $d$ hops from $n$.

\section{Background \& Related work}
\label{sec:bgrw}

Our goal is to build compact multi-view behavior profiles of \apk{ }files in a scalable manner. To this end, we develop a novel \apk{ }embedding framework by combining several RL ideas such as word, document and graph embedding models and feature hashing. Hence, in this section, the related background from these areas are reviewed.

\subsection {Skipgram word and document embedding model}
\label{ss:de}
The popular word embedding model \wv{} \cite{w2v} produces word embeddings that capture meaningful syntactic and semantic regularities. To learn word embeddings, \wv{} uses a simple feed-forward neural network architecture called \textit{skipgram}. It exploits the notion of context such that,  given a sequence of  words $ \{w_1, w_2, ... , w_t, ... , w_T \} $, the target word $w_t$ whose representation has to be learnt and the length of the context window $c$, the objective of skipgram model is to maximize the following log-likelihood:
\begin{equation}
\small
\sum_{t=1}^{|\mathcal{T}|} \log \Pr (w_{t-c},...,w_{t+c}  | w_t) \approx \sum_{t=1}^{|\mathcal{T}|} \log \prod_{-c \le j \le c, j \ne 0} \Pr (w_{t+j} | w_t)
\end{equation}
where $ w_{t-c},...,w_{t+c} $ are the context words and $\mathcal{T}$ is the vocabulary of all the words. 
Here, the context and target words are assumed to be independent. Furthermore, the term $ \Pr (w_{t+j} | w_t) $ is defined as:
$\frac {e^{(\vec{w_t} \cdot  \vec{w'}_{t+j})}} {\displaystyle \sum_{w \in \mathcal{T}} e^{(\vec{w_t} \cdot \vec{w})}}$
where $ \vec{w} $ and $ \vec{w'} $ are the input and output embeddings of word $ w $, respectively. In the face of very large $\mathcal{T}$, the posterior probability in eq.(1) could be learnt in an efficient manner using the so-called \textit{negative sampling} technique. 

\textbf{Negative Sampling.}
In each iteration, instead of considering all words in $\mathcal{T}$ a small subset of words that do not appear in the target word's context are selected at random  to update their embeddings. Training this way ensures the following: \textit{if a word $ w_t $ appears in the context of another word $ w_c $, then the embedding of $ w_t $ is closer to that of $ w_c $ compared to any other randomly chosen word from $\mathcal{T}$.}
Once skipgram training converges, semantically similar words are mapped to closer positions in the embedding space revealing that the learnt embeddings preserve semantics. 

Le and Mikolov's \dv \cite{d2v} extends the skipgram model in a straight forward manner to learn representations of arbitrary length word sequences such as sentences, paragraphs and whole documents. Given a set of documents $\mathbb{D} = \{d_1,d_2,...\}$ and a set of words $ c (d_i) = \{ w_1, w_2, ... \}$ sampled from document $d_i \in \mathbb{D}$, \dv{} skipgram learns a $\delta$ dimensional embeddings of the document $d_i$ and each word $w_j \in c(d_i)$.
The model works by considering a word $w_j \in c(d_i)$  to be occurring in the context of document $d_i$ and tries to maximize the following log likelihood:
$\sum_{j=1}^{|c(d_i)|} \log \Pr\ (w_j  | d_i)$
where the probability  $ \Pr (w_j  | d_i)$ is defined as:
$\frac {e^{(\vec{d} \cdot \vec{w_{j}})}} { \sum_{w \in \mathcal{T}} e^{(\vec{d} \cdot \vec{w})}}$
Here, $\mathcal{T}$ is the vocabulary of all the words across all documents in $\mathbb{D}$.
Understandably, \dv{} skipgram could be trained efficiently using negative sampling.

\textbf{Model parameters.} From the explanations above, it is evident that the total number of trainable parameters of skipgram word and document embedding skipgram models would be $2|\mathcal{T}|\delta$ and $\delta(|\mathbb{D}|+|\mathcal{T}|)$, respectively.

\subsection{Hash embedding model}
\label{ss:bg-he}


Though skipgram emerged as a hugely successful embedding model, it poses scalability issues when the vocabulary $\mathcal{T}$ is very large. Also, its architecture does not support learning embeddings when new words (\textit{aka} \textit{new tokens}) stream over time.  To address these issues, Svenstrup \textit{et al}., \cite{he} proposed a feature hashing based word embedding model. This model involves the following steps:\\
(1) A token to id mapping function, $\mathcal{F}: \mathcal{T} \rightarrow \{1,...,K\}$ and $k$ hash functions of the form $\mathcal{H}_i: \{1,...,K\} \rightarrow \{1,...,B\}, i \in [1,k]$ are defined ($B$ is the number of hash buckets and $B<<K$).\\ 
(2) The following arrays are initialized: $\Phi^{B \times \delta}$: a pool of $B$ component vectors which are intended to be shared by all words in $\mathcal{T}$, and $p^{K \times k}$: contains the importance of each component vector for each word.\\
(3) Given a word $w \in \mathcal{T}$, hash functions $ \mathcal{H}_1,...\mathcal{H}_k $ are used to choose $ k $ component vectors from the shared pool $\Phi$.\\
(4) The component vectors from step (3) are combined as a weighted sum to obtain the hash embedding of the word $w$: $\vec{w} =  \sum_{w,i=1}^{i=k} {p_w^i}\mathcal{H}_i(w)$. \\
(5) With hash embeddings of target and context words thus obtained, skipgram model could be used to train for eq. (1). However, unlike regular skipgram which considers $\Phi$ alone as a set of trainable parameters, one could train $p$ as well. 

Thus Svenstrup et al's framework reduces the number of trainable parameters from $2K\delta$ to $2(B\delta + Kk)$, which helps reducing the pretraining time and memory requirements. The effect of collisions from $K$ to $B$ could be minimized by having more than one hash function and this helps in maintaining accuracies on-par with \wv.
Besides, when new words arrive over time, a function like MD5 or SHA1 could be used in place of $\mathcal{F}$ to hash them to a fixed set of integers in range $ [1,K] $. This helps learning word embeddings in an online fashion.

\subsection{Graph embedding models}
\label{ss:ge}
Analogously, \gv \cite{g2v} considers simple node labeled graphs such as CFGs as documents and the rooted \sg{s} around every node in them as words that compose the document. The intuition is that different subgraphs compose graphs in a similar way that different words compose documents. In this way, \gv{} could be perceived as an RL variant of Weisfeiler-Lehman kernel (\wlk) which counts the number of common rooted \sg s across a pair of graphs to estimate their similarity. As such, \gv{} is capable of learning embeddings of arbitrary sized graphs. 



Given a dataset of graphs $\mathbb{G} = \{G_1, G_2, ...\}$, \gv{} extends the skipgram model explained in \S \ref{ss:de} to learn embeddings of each graph. Let $G_i \in \mathbb{G}$ be denoted as $(N_i,E_i,\lambda)$ and the set of all rooted \sg s around every node $n \in N_i$ (up to a certain degree $D$) be denoted as $c(G_i)$. \gv{} aims to learn a $\delta$ dimensional embeddings of the graph $G_i$ and each subgraph $sg_j$ sampled from $c(G_i)$ i.e., $\vec{G_i},\vec{sg_j} \in \mathbb{R}^{\delta}$, respectively by maximizing the following log likelihood: $
\sum_{j=1}^{|c(G_i)|}\log \Pr\ (sg_j|G_i)$, 
where the probability $\Pr\ (sg_j|G_i)$ is defined as:$\frac{e^{(\vec{G}\cdot{\vec{sg_j}})}}{  \sum_{w\in{\mathcal{T}}}e^{(\vec{G}\cdot{\vec{sg}})}}$.
Here, $\mathcal{T}$ is the vocabulary of all the \sg s across all graphs in $\mathbb{G}$.
The number of trainable parameters of this model will be $\delta(|\mathbb{G}|+|\mathcal{T}|)$. 

Similar to \gv, many recent approaches such as \sv{} \cite{sub2vec}, \gefsg{} \cite{gefsg} and Anonymous Walk Embeddings (AWE) \cite{awe} have adopted skipgram architecture to learn unsupervised graph embeddings. The fundamental difference among them is the type of graph \sss{} that they consider as a graph's context. For instance, \sv{} considers nodes, \gefsg{} considers frequent subgraphs (FSGs) and AWE considers walks that exist in graphs as their contexts, respectively.

\subsection{Semi-supervised embedding model}
Recently, Pan \textit{et al}. \cite{tridnr} extended the skipgram model to learn embedding of nodes in Heterogeneous Information Networks (HINs) in a semi-supervised fashion. Specifically, their extension facilitates skipgram to use class labels of (a subset of) samples while building embeddings. For instance, when $l_i$, the class label of a document $d_i$ is available, the \dv{} model could maximize the following log likelihood:
\begin{equation}
\beta \sum_{j=1}^{|c(d_i)|}\log \Pr\ (w_j|d_i) + (1-\beta) \sum_{j=1}^{|c(d_i)|}\log \Pr\ (w_j|l_i)
\end{equation} 
to include the supervision signal available from $l_i$ along with the contents of the document made available through $c(d_i)$.
Here, $\beta$ is the weight that balances the importance of the two components in document embedding. Pan \textit{et al}. empirically prove that embeddings with this form of semi-supervision significantly improves the accuracy of downstream tasks.

In summary, all above-mentioned embedding models possess some strengths for RL in various areas mainly for natural language texts.  Differing from them, we propose a new and efficient data-driven graph embedding model/framework for app behaviour profiling. The new framework has three unique characteristics which brings in crucial strengths for holistic app behavior profiling, namely: (i) multi-view RL, (ii) semi-supervised RL, and (iii) feature hashing based RL.  We illustrate (both in principle and through experiments) that this new framework possess all these strengths and caters a multitude of downstream tasks.

\section{Methodology}
\label{sec:meth}
In this section, we explain the \tool{} app profiling framework. We first present an overview of the framework which encompasses two phases, subsequently, we discuss the details of each component in the following subsections.

\subsection{Framework Overview}
\begin{figure}[t]
	\includegraphics[height=5.5cm,width=9cm]{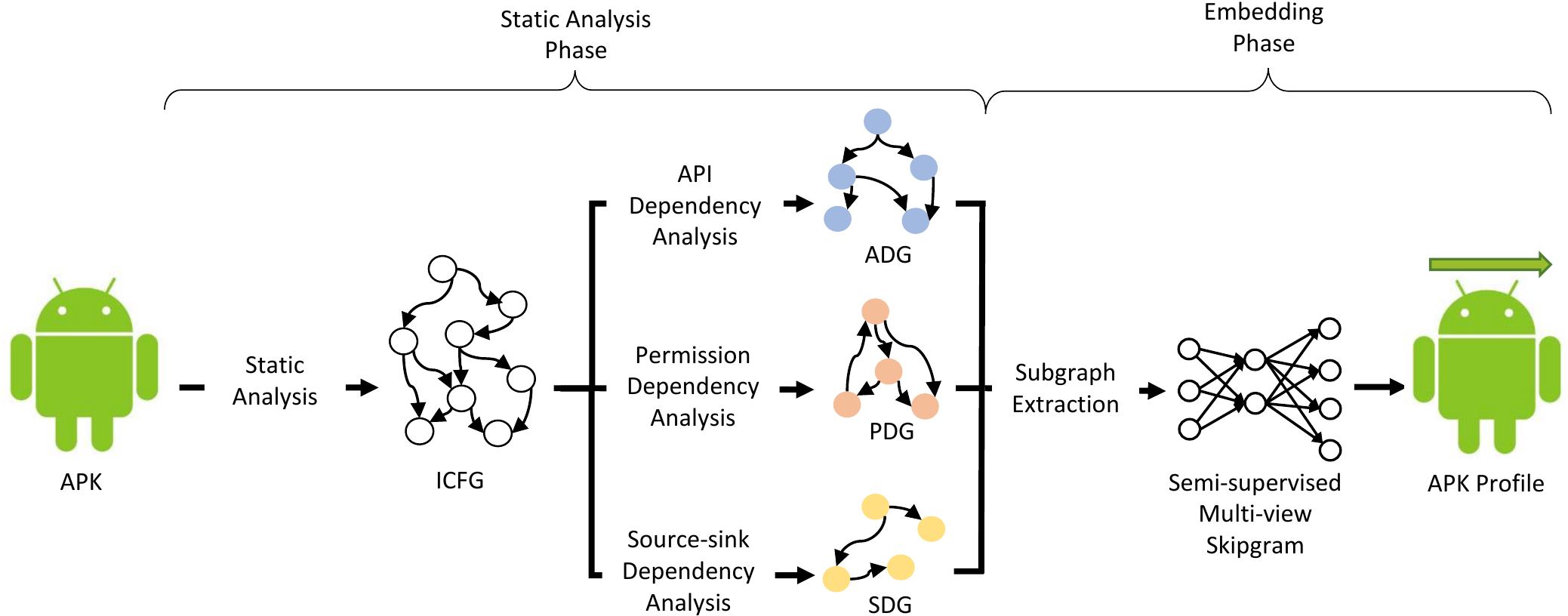}
	\caption{APK2VEC: Framework overview 
		\label {fig:ov}}
\end{figure}

As depicted in Figure \ref{fig:ov}, the workflow of \tool{} can be divided into two major phases, namely, the \textit{static analysis phase} and the \textit{embedding phase}. The static analysis phase encompasses of only program analysis procedures which intend to transform \apk{ }files into DGs. The subsequent embedding phase encompasses RL techniques that transform these DGs into \apk{ }profiles. 

\begin{figure}[t]
	\centering
	\includegraphics[height=6cm,width=7cm]{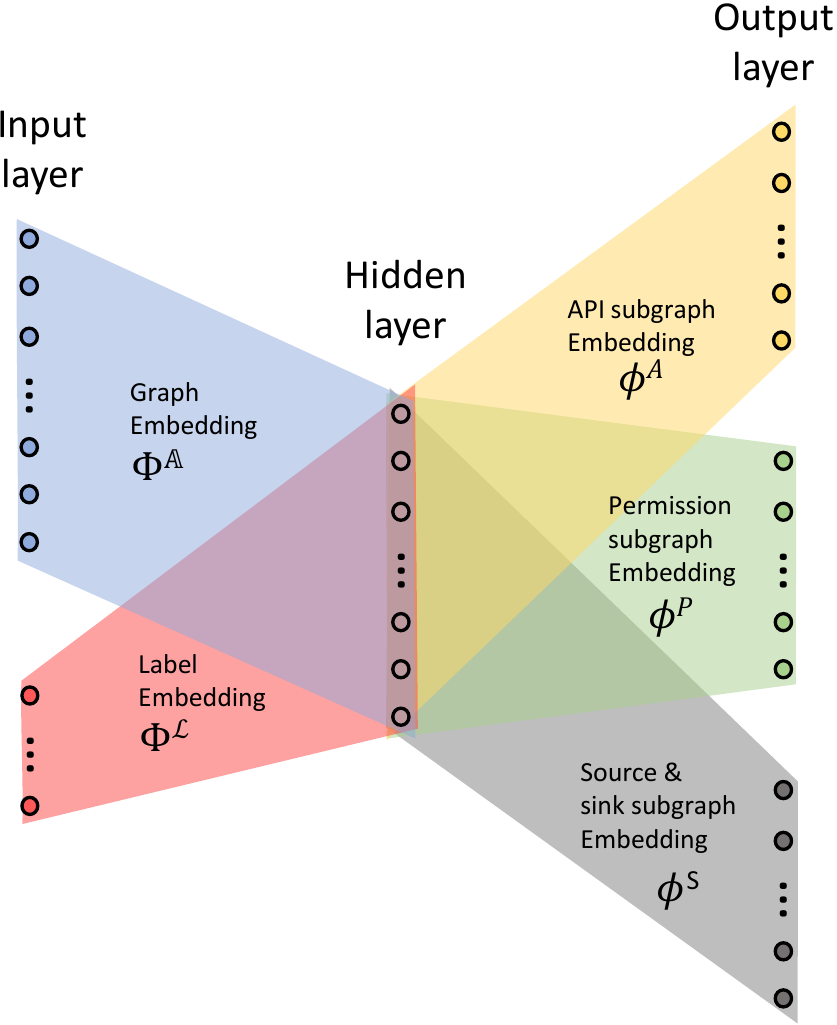}
	\caption{Semi-supervised Multi-view Skpigram 
		\label {fig:sgns}}
\end{figure}
\textbf{Static analysis phase.} This phase begins with disassembling the \apk{s} in the given dataset  and constructing their interprocedural CFG (ICFG). Further static analysis is performed to abstract each ICFG into three different DGs, namely, ADG, PDG and SGD. Each of them represent a unique semantic view of the app's behaviors with distinct modalities. Detailed procedure of constructing these DGs is presented in \S \ref{ss:sa}.

\textbf{Embedding phase.} After obtaining the DGs for all the \apk s in the dataset, rooted \sg{s} around every node in the DGs are extracted to facilitate the learning  \apk{ }embeddings. Once the rooted \sg{s} are extracted, we train the semi-supervised multi-view skpigram neural network with them. The detailed procedure is explained in \S \ref{ss:ep}.

\subsection{Static Analysis Phase}
\label{ss:sa}

\textbf{ICFG construction.} Given an \apk{ }file, the first step is to perform static control-flow analysis and construct its ICFG. We chose ICFG over other program representation graphs (e.g., DFGs, call graphs) due to its fine-grained representation of control flow sequence, which allows us to capture finer semantic details of the \apk{} which is necessary to build a comprehensive \apk{} profile. Formally, $ICFG = (N, E)$ for an \apk{ }$a$ is a directed graph where each node $bb \in N$ denotes a basic block\footnote{ A basic block is a sequence of instructions in a method with only one entry point and one exit point which represents the largest piece of the program that is always executed altogether.} of a method $m$ in $a$, and each edge $e(bb_1, bb_2) \in E$ denotes either an intra-procedural control-flow or a calling relationship from  $bb_1$ to $bb_2$ and $E \subseteq N \times N$.

\textbf{Abstraction into multiple views.} Having constructed the ICFG, to obtain  richer semantics from the \apk{}, we abstract it with three Android platform specific analysis, namely API sequences, Android permissions, and information sources \& sinks to construct the three DGs, respectively. The abstraction process is described below. 

To obtain the ADG from a given ICFG, we remove all nodes that do not access security sensitive Android APIs. This will leave us with a subset of sensitive nodes from the perspective of API usages, say $N^\mathcal{A} \subseteq N$. Subsequently, we connect a pair of nodes $n_1,n_2 \in N^\mathcal{A}$ iff there exist a path between them in ICFG. This yields the ADG, $G^\mathcal{A}$ which could be formally represented as a three tuple $G^\mathcal{A} = ( N^\mathcal{A},E^\mathcal{A},\lambda^\mathcal{A})$, where $\lambda^\mathcal{A}:N^\mathcal{A}\rightarrow L^\mathcal{A}$ is a labeling function that assigns a security sensitive API as a node label to every node in $N^\mathcal{A}$ from a set of alphabets $L^\mathcal{A}$. We refer to existing work \cite{casandra} for the list of security sensitive APIs. Similarly, we use works such as PScout \cite{pscout} and SUSI \cite{susi} which maps APIs to Android permissions and information source/sinks to obtain set of node labels $L^\mathcal{P}$ and $L^\mathcal{S}$, respectively. Subsequently, adopting the process mentioned above using them we abstract the ICFG into PDG and SDG using $L^\mathcal{P}$ and $L^\mathcal{S}$, respectively.

\subsection{Embedding Phase}
\label{ss:ep}

In the embedding phase, our goal is to take the DGs and class labels that correspond to a set of \apk{s} and train the skipgram model to obtain the behavior profile for each  \apk. To this end, we develop a novel variant of the skipgram model which facilitates incorporating the three following learning paradigms: semi-supervised RL, multi-view RL and feature hashing.

\textbf{Network architecture}
Figure \ref{fig:sgns} depicts the architecture of the neural network used in \tool's RL process. The network consists of two shared input layers and three shared output layers (one for each view). The goal of the first input layer ($\Phi^\mathbb{A}$) is to perform multi-view RL. More precisely, given an \textit{apk} id $a_i$ in the first layer, the network intends to predict the API, permission and source-sink subgraphs that appear in $a_i$'s context, in each of the output layers. Similarly, the goal of the second input layer ($\Phi^\mathcal{L}$) is to perform semi-supervised RL. More specifically, given $a_i$'s class label as input in the second layer, the network intends to predict subgraphs of all three views that occur in $a_i$'s context in each of the output layers. Thus, the network forces API, permission and source-sink subgraphs that frequently co-occur with same class labels to have similar embeddings. For instance, given a malware family label $\mathfrak{f}$, subgraphs that characterize $\mathfrak{f}$'s behaviors would end up having similar embeddings. This in turn would influence \textit{apk}s that belong to $\mathfrak{f}$ to have similar embeddings. 

\textbf{Hash embeddings.} Considering the real-world scenario where Android platform evolves (i.e., APIs/permissions being added or removed) and apps stream rapidly over time, it is obvious that the vocabulary of subgraphs (across all DGs) would grow as well. Regular skipgram models could not handle such a vocabulary and as mentioned in \S \ref{ss:bg-he}, hash embeddings could be used effectively to address this. Note that in our framework, the vocabulary of tokens is only present in the output layer. Hence, in \tool, hash embeddings are used only in the three output layers ($\phi^A, \phi^P$ and $\phi^S$) and the two input layers ($\Phi^\mathbb{A}$ and $\Phi^\mathcal{L}$) uses regular embeddings.

The process through with our skipgram model is trained is explained through Algorithm  \ref{algo:apk2vec}.

\begin{algorithm}[t]
	\scriptsize
	\LinesNumbered
	\caption{ \textsc{apk2vec} ($ \mathbb{A}, \mathbb{L}, D, \delta, \mathcal{E}, B^v, k, \alpha$) \label{algo:apk2vec}}
	\KwIn{$\mathbb{A} = \{a_1, a_2, ...\}$: set of \apk{s} such that $a_i = \{G_i^{v}\}$ for $v \in \{\mathcal{A}, \mathcal{P}, \mathcal{S}\}$\newline
		$\mathbb{L} =\{l_1,l_2,...\}:$ Set of labels for each \apk{ }in $\mathbb{A}$. Note that there may be zero or more labels for an \apk. Hence $\forall l_i \in \mathbb{L}, |l_i| \ge 0$. Let the total number of unique labels across $l_i \ \in \mathbb{L}$ be denoted as $\mathcal{L}$.\newline
		$D:$ Maximum degree of rooted subgraphs to be considered for learning embeddings. This will produce a vocabulary of subgraphs in each view, $\mathcal{T}^{v}$ = $\{sg_1^{v}, sg_2^{v}, ...\}$ from all the graphs $G_{i}^{v}$. Let $|\mathcal{T}^v|$ be denoted as $K^{v}$. \newline
		$\delta:$ Number of dimensions (embedding size)\newline
		$\mathcal{E}:$ Number of epochs\newline
		$B^{v}$: Number of hash buckets for view {$v$}\newline
		$k$: Number of hash functions  (maintained same across all views) \newline
		$\alpha:$ Learning rate}
	\KwOut{Matrix of vector representations of \apk{s} $\Phi^\mathbb{A} \in \mathbb{R}^{|\mathbb{A}| \times \delta}$ }
	Initialization: Sample $\Phi^\mathbb{A}$ from $\mathbb{R}^{\mathbb{|{A}|} \times \delta}$, $\Phi^\mathcal{L}$ from $\mathbb{R}^{|\mathcal{L}| \times \delta}$, $\phi^{v}$ from $\mathbb{R}^{B^v \times k}$, and $p^{v}$ from $\mathbb{R}^{K^v \times k}$ \newline
	\For{$e = \{1, 2, ..., \mathcal{E} \}$}{
		\For{ $a_i \in \textsc{Shuffle}(\mathbb{A})$}{
			\For{$G_i^{v} \in a_i$}{
				$sg_c := \textsc{GetSubgraphs} (G_i^{v},D)$ \\
				$J(\Phi^\mathbb{A},\phi^{v},p^{v})$ := $-\ \log \displaystyle \prod_{sg \in sg_c} \frac{e^{(\Phi^\mathbb{A}(a_i) \cdot \textsc{HashEmb}({sg},\phi^{v},p^{v},v))}}{\displaystyle \sum_{sg' \in \mathcal{T}^{v}} e^{(\Phi^\mathbb{A}(a_i) \cdot \textsc{HashEmb}({sg',\phi^{v},p^{v},v}))}} $ \\
				$ \Phi^\mathbb{A} := \Phi^\mathbb{A} - \alpha  \frac{\partial J}{\partial \Phi^\mathbb{A}} $;
				$ \phi^{v} := \phi^{v} - \alpha  \frac{\partial J}{\partial \phi^{v}} $;
				$ p^{v} := p^{v} - \alpha  \frac{\partial J}{\partial p^{v}} $
			}	
			\For{$l \in l_i$}{
				$J(\Phi^\mathcal{L},\phi^{v},p^{v})$	:= $-\ \log \displaystyle \prod_{sg \in sg_c} \frac{e^{(\Phi^\mathcal{L}(l) \cdot \textsc{HashEmb}({sg},\phi^{v},p^{v},v))}}{\displaystyle \sum_{sg' \in \mathcal{T}^{v}} e^{(\Phi^\mathcal{L}(l) \cdot \textsc{HashEmb}({sg',\phi^{v},p^{v},v}))}} $ \\
				$ \Phi^\mathcal{L} := \Phi^\mathcal{L} - \alpha  \frac{\partial J}{\partial \Phi^\mathcal{L}} $;
				$ \phi^{v} := \phi^{v} - \alpha  \frac{\partial J}{\partial \phi^{v}} $;
				$ p^{v} := p^{v} - \alpha  \frac{\partial J}{\partial p^{v}} $
			}
		}
		\textbf{return} $\Phi^\mathbb{A}$}
\end{algorithm}

\subsection{Algorithm: \tool}
The algorithm takes the set of \apk{s} along with their corresponding DGs ($\mathbb{A}$), set of their labels ($\mathbb{L}$), maximum degree of rooted subgraphs to be considered ($ D $), embedding size ($ \delta $), number of epochs ($ \mathcal{E} $), number of hash buckets per view ($ B^v $), number of hash functions ($k$) and learning rate ($ \alpha $) as inputs and outputs the \apk{ }embeddings ($ \Phi^{\mathbb{A}} $). The major steps of the algorithm are as follows:
\begin{enumerate}
	[leftmargin=*]
	\setlength\itemsep{0em}
	\item We begin by randomly initializing the parameters of the model i.e., $\Phi^\mathbb{A}$: \apk{ }embeddings, $\Phi^\mathcal{L}$: label embeddings, $\phi^v$: embeddings of each hash bucket for each of the three views, and $p^v$: importance parameters for each of the views (line 1). It is noted that except the \apk{ }embeddings, all other parameters are discarded when training culminates.
	
	\item For each epoch, we consider each \apk{ }$a_i$ as the target whose embedding has to be updated. To this end, each of its DG $G^v_i$ is taken and all the rooted subgraphs upto degree $D$ around every node are extracted from the same (line 4). The \sg{} extraction process is explained in detail in \S \ref{ss:sgex}.
	
	\item The set of all such subgraphs $sg_c$, is perceived as the context of $a_i$. Once $sg_c$ is obtained, we get their hash embeddings and compute the negative log likelihood of them being similar to the target \apk{ }$a_i$'s embedding (line 5). The hash embedding computation process is explained in detail in \S \ref{ss:he}. 
	
	\item With the loss value thus computed, the parameters that influence the loss are updated (line 6). We propose a novel view-specific negative sampling strategy to train the skipgram and the same is explained in \S \ref{ss:ns}.
	
	\item Subsequently, for each of $a_i$'s class labels i.e., $l \in l_i$, we compute the negative log likelihood of their similarity to the context subgraphs in $sg_c$ and update the parameters that influence the same (lines 7-9). This step amounts to semi-supervised RL as $l_i$ could be empty for some \apk{s}.
	
	\item The above mentioned process is repeated for $\mathcal{E}$ epochs and the \apk{ }embeddings (along with other parameters) are refined. 
\end{enumerate}

Finally, when training culminates, \apk{ }embeddings in $\Phi^\mathbb{A}$ are returned (line 10).

\subsection {Extracting context subgraphs}
\label{ss:sgex}
For a given \apk{ }$a_i$, extracting rooted \sg s around each node in each $G_i^v$ and considering them as its context is a fundamental task in our approach. To extract these \sg s, we follow the well-known Weisfeiler-Lehman relabeling process \cite{wlk} which lays the basis for \wlk{}  \cite{dgk,wlk}. The subgraph extraction process is presented formally in Algorithm \ref{algo:get_rooted_sg}. The algorithm takes the graph $ G $ from which the subgraphs have to be extracted and maximum degree to be considered around root node $ D $ as inputs and returns the set of all rooted subgraphs in $G$, $ S $. It begins by initializing $S$ to an empty set (line 2). Then, we intend to extract rooted subgraph of degree $d$ around each node $n$ in the graph. When $ d =0 $, no subgraph needs to be extracted and hence the label of node $ n $ is returned (line 6). For cases where $ d > 0 $, we get all the (breadth-first) neighbours of $ n $ in $ \mathcal{N}_n $ (line 8). Then for each neighbouring node, $ n' $, we get its degree $ d-1 $ subgraph and save the same in a multiset $ M^{(d)}_n $ (line 9). Subsequently, we get the degree $ d-1 $ subgraph around the root node $ n $ and concatenate the same with sorted list $ M_n^{(d)} $ to obtain the subgraph of degree $d$ around node $n$, which is denoted as $ sg_n^{(d)} $ (line 10). $ sg_n^{(d)} $ is then added to the set of all subgraphs (line 11). When all the nodes are processed, rooted subgraphs of degrees $[0,D]$ are collected in $S$ which is returned finally (line 12). 

\begin{algorithm}[t]
	\scriptsize
	\LinesNumbered
	\caption{\textsc{GetSubgraphs} $ (G, D) $ \label{algo:get_rooted_sg}}
	
	\Begin{
		$S := \{\}$ \textcolor{gray}{//Initialize with an empty set}\\
		\For{ $n \in N$ }{
			\For {$d \in \{0,1,...,D\}$}{
				\If {$d = 0$} {
					$sg^{(d)}_n := \lambda(n) $ \textcolor{gray}{//node label}
				}
				\Else{
					$\mathcal{N}_n := \{n'\ |\ (n,n') \in E\}$\textcolor{gray}{//neighboring nodes}\\
					$M^{(d)}_n := \{\{$\textsc{GetWLSubgraph}$(n',G,d-1)\ |\ n' \in \mathcal{N}_n \}\}$ \textcolor{gray}{//multiset of rooted subgraphs around neighboring nodes}\\
					$sg^{(d)}_n := \textsc{GetWLSubgraph} (n,G,d-1) \oplus sort(M^{(d)}_n)$  
				}
				$S := S \cup sg^{(d)}_n$
			}
		}
		
		\textbf{return } $ S $ \textcolor{gray}{//set of all rooted subgraphs in $G$}\\
	}
\end{algorithm}

\subsection{Obtaining hash embeddings}
\label{ss:he}

\begin{algorithm}[t]
	\scriptsize
	\LinesNumbered
	\caption{ $\textsc{HashEmb}$ ($sg,\phi,p,v$) 
		\label{algo:he}}

	\Begin{
		$sg_{id}$ := $\mathcal{F}^v (sg)$ \textcolor{gray}{//Token to id mapping function}\\
		$components := (\phi(\mathcal{H}_1(sg_{id})),...,\phi(\mathcal{H}_k(sg_{id}))^T $\textcolor{gray}{//shape of components: $k \times \delta$}\\
		$weights := (p^v_1(sg_{id}),p^v_2(sg_{id}),...,p^v_k(sg_{id}))^T$\textcolor{gray}{//shape of weights: $k \times 1$}\\ 
		$ \vec{sg} := weights^T \cdot components $ \textcolor{gray}{//$ 1 \times k \cdot k \times \delta $}\\
		\textbf{return} $\vec{sg}$
	}	
\end{algorithm}

Once context subgraphs are extracted, we proceed with obtaining their hash embeddings and training the same along the target \apk{}'s embedding. 
Given a \sg, the process of extracting its hash embedding involves four steps which are formally presented in Algorithm \ref{algo:he}. Following is the explanation of this algorithm:
\begin{enumerate}
	[leftmargin=*]
	\setlength\itemsep{0em}
	\item Given a \sg{} $sg$, we begin by mapping to an integer $sg_{id}$, using a function $\mathcal{F}^v$ (line 2). When $\mathcal{T}^v$, the vocabulary of all the \sg s in view $v$ could be obtained ahead of training, a regular dictionary \textit{aka} token-to-id  function which maps each \sg{} to a unique number in the range $[1,K^v]$ (where $K^v = |\mathcal{T}^v|$) could be used as $\mathcal{F}^v$. In the online learning setting, such a dictionary could not be obtained. Hence, analogous to feature hashing \cite{hashtrick}, one could use a regular hash function such as MD5 or SHA1 to hash the subgraph to an integer in the predetermined range $[1,K^v]$ (here, an arbitrarily large value of $K^v$ is chosen to avoid collisions).
	
	\item $sg_{id}$ is then hashed using each of the $k$ hash functions. Each function $\mathcal{H}_i, i \in [1,k]$ maps it to one of the $B^v$ available hash buckets which in turn maps to one of the $B^v$ component embeddings in $\phi^v$. Thus we obtain $k$ component embeddings for the given subgraph and save them in $components$ (line 3). In other words, $components$ contains $k$ $\delta$-dimensional embeddings.
	
	\item Similarly, using $sg_{id}$, we then lookup the importance parameter for each hash function, $p^v_i (sg_{id}), i \in [1,k]$ and save them in $weights$ (line 4). In other words, $weights$ contains $k$ importance values.
	
	\item Finally, the hash embedding of the \sg{} is obtained by multiplying $k$ $\delta$-dimensional component vectors with $k$ corresponding importance values (line 5).
\end{enumerate} 

Once the hash embeddings of the context \sg s are obtained using the above mentioned process, one could train them along with the target \apk{}'s embedding using a learning algorithm such as Stochastic Gradient Decent (SGD).

\subsection{View-specific negative sampling}
\label{ss:ns}

Similar to other skipgram based embedding models such as \gv{} \cite{g2v}, we could efficiently minimize the negative log likelihood in lines 5 and 8 of Algorithm \ref{algo:apk2vec}. That is, given an \apk{ }$a$ and a subgraph $sg^v$ which is contained in view $v$, the regular negative sampling intends to maximize the similarity between their embeddings. Besides, it chooses $\eta$ subgraphs as negative samples i.e., that do not occur in the context of $a$ and minimizes the similarity of $a$ and these negative samples. This could be formally presented as follows,
\begin{equation}
\Pr (sg^v|a) = \sigma(\vec{a}^T\cdot\vec{sg^v})\prod_{j=1}^\eta\mathbb{E}_{sg_j \sim \Pr_n(\mathcal{T})} \sigma(-\vec{a}^T\cdot\vec{sg_j})
\end{equation}
where, $\mathcal{T} = \bigcup_v \mathcal{T}^v$ is union of vocabularies across all views and $\mathbb{E}$ is expectation of choosing a subgraph $sg_j$ from the smoothed distribution of subgraphs $\Pr_n$ across all the three views. 

In simpler terms, eq. (3) moves $\vec{a}$ closer to $\vec{sg^v}$ as it occurs in $a$'s context and also moves $\vec{a}$ farther away from $\vec{sg_j}$ (which may not belong to view $v$) as it does not occur in $a$'s context.

However, in our multi-view embedding scenario, the distribution of subgraphs is not similar across all views. For instance, in our experiments reported in \S \ref {sec:eval}, the API view produces millions of subgraphs, where as the permission and source-sink view produce only thousands. Hence, eq. (3) which ignores the view-specific probability of subgraph occurrences is not suitable in this scenario. Therefore, we propose a novel view-specific negative sampling strategy as described by the equation below:
\begin{equation}
\Pr (sg^v|a) = \sigma(\vec{a}^T\cdot\vec{sg^v})\prod_{j=1}^\eta\mathbb{E}_{sg^v_j \sim \Pr_n( \mathcal{T}^v)} \sigma(-\vec{a}^T\cdot\vec{sg^v_j})
\end{equation}

In simpler terms, eq. (4) moves $\vec{a}$ closer to $\vec{sg^v}$ as it occurs in $a$'s context and also moves $\vec{a}$ farther away from $\vec{sg_j^v}$ (which also belongs to view $v$) as it does not occur its $a$'s context.

\subsection{Model dynamics}
\label{ss:md}
{\color{black} The trainable parameters of our model are $\Phi^\mathbb{A} , \Phi^\mathcal{L}, \phi^v$, and $p^v$. Recall, $\Phi^\mathbb{A}$ and $\Phi^\mathcal{L}$ are regular embeddings as they are in the input layers and $\phi^v$s  are hash embeddings.
	Also, the total number of tokens in the input and output layers would be $|\mathbb{A}| + |\mathcal{L}|$ and $\sum_v K^v$, respectively. 
	Hashing (which is applicable only to $\phi^v$) reduces the number of parameters in the output layer from $\sum_v K^v$ to $K^v  k + k  B^v$ where $B^v << K^v$ (typically, we set $k=[2,4]$ and $K^v > B^v \cdot 100$).
	
	From the explanations above, it is evident that the computational overhead of using hash embeddings instead of standard embeddings is in the embedding lookup step. More precisely, a multiplication of a $1 \times k$ matrix (obtained from $p^v$) with a $k \times \delta$ matrix (obtained from $\phi^v$) is required instead of a regular matrix lookup to get $1 \times \delta$ \sg{} embedding. When using small values of $k$, the computational overhead is therefore negligible. In our experiments, hash embeddings are marginally slower to train than standard embeddings on datasets with small vocabularies. 


	\section{Evaluation}
	\label{sec:eval}
	
	We evaluate the efficacy of \tool's embeddings with several tasks involving various learning paradigms that include supervised learning (batch and online), unsupervised learning and link prediction. The evaluation is carried out on five different datasets involving a total of 42,542 Android apps. In this section, we first present the experimental design aspects, such as research questions addressed, datasets and tasks chosen pertaining to the evaluation. Subsequently, the results and relevant discussions are presented.
	
	\textbf{Research Questions.}
	Through our evaluations, we intend to address the following questions:
	\begin{itemize}
		[leftmargin=*]
		\setlength\itemsep{0em}
		\item {How accurate do \tool's embeddings perform on various app analytics tasks and how do they compare to \soa{} approaches?}
		\item {Do multi-view profiles offer better accuracies than single-view profiles?}
		\item {Does semi-supervised RL help improving the accuracy of app profiles?}
		\item {How does \tool's hyperparameters affect its accuracy and efficiency?}
	\end{itemize}
	\textbf{Evaluation setup.} All the experiments were conducted on a server with 40 CPU cores (Intel Xeon(R) E5-2640 2.40GHz), 6 NVIDIA Tesla V100 GPU cards with 256 GB RAM running Ubuntu 16.04.
	
	\textbf{Comparative analysis.} To provide a comprehensive evaluation, we compare our approach with four baseline approaches, namely,  \wlk{} \cite{wlk}, \gv{} \cite{g2v}, \sv \cite{sub2vec} and \gefsg{} \cite{gefsg}. Refer to \S \ref{sec:intro} and \S \ref{sec:bgrw} for brief explanations on the baselines. The following evaluation-specific details on baselines are noted: (i) Since all baselines are unimodal they are incapable of leveraging all the three DGs to yield one unified \apk{} embedding. Hence, to ensure fair comparison, we merge all three DGs into one graph and feed them to these approaches. (ii) \sv{} has two variants, namely, \svn{} (which leverages only neighborhood information for graph embedding) and \svs{} (which leverages only structural information). Both these variants are included in our evaluations, and (iii) For all baselines except \gefsg{}, open-source implementations provided by the authors are used. For \gefsg{}, we reimplemented it by following the process described in their original work. Our reimplementation could be considered faithful as it reproduces the results reported in the original work.
	
	\textbf{Hyperparameter choices.}
	In terms of \tool's hyperparameters, we set the following values: $\mathcal{E}=100, \delta=64$, $\alpha=0.1$ (with decay) and $\eta=2$. When hash embedding is used $k=2, B^v=\frac{K^v}{10}$ (for all $v$). To ensure fair comparison, in all experiments, the hyperparameters of all baseline approaches are maintained same as those of \tool{} (e.g., the embedding dimensions of all baselines are set to 64, etc.). In all experiments, for datasets where class labels are available, 25\% of the labels are used for semi-supervision during embedding (unless otherwise specified). 
	
	\begin{table}[]
		\centering

		\caption{Datasets used for evaluations}
		\label{tab:rq1-exp}
		\resizebox{\columnwidth}{!}{%
			\begin{tabular}{|c|c|c|c|c|c|c|c|c|}
				\hline
				\multirow{2}{*}{\textbf{Task}} & \multirow{2}{*}{\textbf{Data source}} & \multirow{2}{*}{\textbf{\# of apps}} & \multicolumn{3}{c|}{\textbf{Avg. nodes}} & \multicolumn{3}{c|}{\textbf{Avg. edges}} \\ \cline{4-9} 
				&  &  & \textbf{ADG} & \textbf{PDG} & \textbf{SDG} & \textbf{ADG} & \textbf{PDG} & \textbf{SDG} \\ \hline \hline
				\multirow{2}{*}{\begin{tabular}[c]{@{}c@{}}Batch malware \\ detection\end{tabular}} & Malware: \cite{drebin,vs} & 19,944 & \multirow{2}{*}{783.03} & \multirow{2}{*}{131.73} & \multirow{2}{*}{80.12} & \multirow{2}{*}{2604.28} & \multirow{2}{*}{174.64} & \multirow{2}{*}{94.04} \\ 
				& Benignware: \cite{gp} & 20,000 &  &  &  &  &  &  \\ \hline
				\multirow{2}{*}{\begin{tabular}[c]{@{}c@{}}Online malware \\ detection\end{tabular}} & Malware: \cite{drebin} & 5,560 & \multirow{2}{*}{365.18} & \multirow{2}{*}{63.11} & \multirow{2}{*}{37.885} & \multirow{2}{*}{744.99} & \multirow{2}{*}{66.96} & \multirow{2}{*}{34.67} \\ 
				& Benignware: \cite{gp} & 5,000 &  &  &  &  &  &  \\ \hline
				\begin{tabular}[c]{@{}c@{}}Malware familial \\ clustering\end{tabular} & Drebin \cite{drebin} & 5,560 & 229.82 & 69.96 & 43.41 & 464.73 & 72.34 & 44.57 \\ \hline
				Clone detection & Clone apps  \cite{kaiclones} & 280 & 674.71 & 179.09 & 94.69 & 1553.29 & 182.24 & 76.64 \\ \hline
				\begin{tabular}[c]{@{}c@{}}App \\ recommendation\end{tabular} & Googleplay \cite{gp} & 2,318 & 2168.88 & 242.87 & 154.14 & 5137.47 & 348.67 & 150.87 \\ \hline
			\end{tabular}%
		}
	\end{table}

	\subsection{\tool{} vs. \soa}
	\label{ss:rq1}
	
	In the following subsections we intend to evaluate \tool{} against the baselines on two classification (i.e., batch and online malware detection), two clustering (i.e., app clone detection and malware familial clustering) and one link prediction (i.e., app recommendation) tasks. To this end, the datasets reported in Table \ref{tab:rq1-exp} are used. It is noted that, DGs used in our experiments are much larger than benchmark datasets (e.g., datasets used in \cite{gefsg}) and even some large real-world datasets (e.g., used in \cite{dgk}). It is noted that some baselines do not scale well to embed such large graphs and they run into \textit{Out of Memory} (OOM) situations. 
	
	\subsubsection{Graph classification} \hfill \\
	\label{ss:rq1.1}
	\textbf{Dataset \& experiments.} For batch learning based malware detection task, 19,944 malware from two well-known malware datasets \cite{drebin} and \cite{vs} are used. To form the benign portion, 20,000 apps from Google Play \cite{gp} have been used. To perform detection, we first obtain profiles of all these apps using \tool{}. Subsequently, a Support Vector Machine (SVM) classifier is trained with 70\% of samples and is evaluated with the remaining 30\% samples (classifier hyperparamters are tuned using 5-fold cross-validation). This trial is repeated 5 times and the results are averaged. 
	
	For online malware detection task, 5,560 malware from \cite{drebin} and 5,000 benign apps from Google Play are used. In this experiment, the real-world situation where apps stream in over time is simulated as follows: First, \apk{s} are temporally sorted according to their time of release (see \cite{casandra} for details). Thereafter, the embeddings of first 1,000 \apk{s} are used to train an online Passive Agressive (PA) classifier. For the remaining 9,560 \apk{s}, their embeddings are obtained in an online fashion using \tool{} as and when they stream in. These embeddings are fed to the trained PA model for evaluation and classifier update. 
	
	For both batch and online settings, to evaluate the efficacy, standard metrics such as precision, recall and f-measure are used. 
	
	\textbf{Results \& discussions.} The batch and online malware detection results are presented in Table \ref{tab:rq1.1}. The following inferences are drawn from the tables.
	\begin{itemize}
		[leftmargin=*]
		\setlength\itemsep{0em}
		\item In batch learning setting, as it is evident from the f-measure, \tool{} outperforms all baselines. More specifically, with just 25\% labels it is able to outperform the worst and best performing baselines by more than 20\% and nearly 2\%, respectively. Clearly, this improvement could be attributed to \tool's multimodal and semi-supervised embedding capabilities.
		
		\item \tool's improvements in f-measure are even more prominent in the online learning setting. More specifically, it outperforms the worst and best performing baselines by nearly 35\% and more than 5\%, respectively. Clearly, this improvement could be attributed to \tool's hash embedding capabilities through which it handles dynamically expanding vocabulary of subgraphs.
		
		\item Looking at the performances of baselines, one could see all of them perform reasonably better in the batch learning setting than the online setting. This is owing to their inability to handle vocabulary expansion which renders their models obsolete over time. Besides, none of them posses multi-view and semi-supervised learning potentials which could explain their overall substandard results.
		
		\item Due to poor space complexity, \gefsg{} \cite{gefsg} is unable to handle the large graphs used in this experiment and went OOM during the FSG extraction process. Hence, its results are not reported in the table.

		
		
		
	\end{itemize}
	
	\begin{table}[]
		\centering
		\scriptsize
		\caption{Malware detection (graph classification) results}
		\label{tab:rq1.1}
		\begin{tabular}{|c|c|c|c|c|c|c|}
			\cline{1-7}
			& \multicolumn{3}{c|}{\textbf{Batch}} & \multicolumn{3}{c|}{\textbf{Online}} \\ \cline{2-7}
			\textbf{Technique} & \textbf{P}(\%) & \textbf{R}(\%) & \textbf{F}(\%)  & \textbf{P}(\%) & \textbf{R}(\%) & \textbf{F}(\%)\\ \hline \hline
			\tool & 88.07 &\textbf{90.41} & \textbf{89.22} &\textbf{87.90}&\textbf{89.73}&\textbf{88.81}\\ 
			\wlk \cite{wlk} &\textbf{88.15} & 86.38 & 87.25 &84.13&82.32&83.22\\ 
			\gv \cite{g2v} & 76.96 & 82.48 & 79.63 &82.55&84.21&83.37\\ 
			\svn \cite{sub2vec} & 68.31 & 69.65 & 68.98 &52.20&56.65&54.33\\ 
			\svs \cite{sub2vec} & 66.94 & 68.36 & 67.64 &53.13&54.98&54.04\\ \hline
		\end{tabular}
	\end{table}
	
	\subsubsection{Graph clustering} \hfill\\
	\label{ss:rq1.2}
	\textbf{Dataset \& experiments.} We now evaluate \tool{} on two different graph clustering tasks. Firstly, in the malware familial clustering task, 5,560 apps from Drebin \cite{drebin} collection are used. These apps belong to 179 malware families. Malware belonging to the same family are semantically similar as they perform similar attacks. Hence, we obtain the profiles of these apps and cluster them into 179 clusters using k-means algorithm. Profiles of samples belonging to same family are expected to end up in the same cluster. 
	
	The next task is clone detection which uses 280 apps from Chen \textit{et al}.'s \cite{kaiclones} work. The apps in this dataset belong to 100 clone groups, where each group contains at least two apps that are semantic clones of each other with slight modifications/enhancements. 
	Hence, in this task, we obtain the app profiles and cluster them into 100 clusters using k-means algorithm with the expectation that cloned 
	apps end up in the same cluster. 
	
	Adjusted Rand Index (ARI) is used as a metric to determine the clustering accuracy in both these tasks. 
	
	\textbf{Results \& discussions.} The clustering results are presented in Table \ref{tab:rq1.2}. The following inferences are drawn from the table.
	\begin{itemize}
		[leftmargin=*]
		\setlength\itemsep{0em}
		\item At the outset, it is clear that \tool{} outperforms all the baselines on both these tasks. For familial clustering and clone detection, the improvements over the best performing baselines are 0.07 and 0.01 ARI, respectively. 
		
		\item Interestingly, unlike malware detection not all the baselines offer agreeable performances in these two tasks. For instance, the ARIs of \sv{} and \gefsg{} are too low to be considered as practically viable solutions. Given this context, \tool's performances show that its embeddings generalize well and are task-agnostic.  
		

	\end{itemize}
	
	\begin{table}[]
		\centering
		\scriptsize
		\caption{Malware familial clustering and clone detection (graph clustering) results}
		\label{tab:rq1.2}
		\resizebox{\columnwidth}{!}{%
			\begin{tabular}{c|c|c|}\hline
				\multicolumn{1}{|c|}{\textbf{Technique}} & \textbf{Familial clustering} (ARI) & \textbf{Clone detection} (ARI) \\ \hline\hline
				\multicolumn{1}{|c|}{\tool} & \textbf{0.5124} & \textbf{0.8360} \\ 
				\multicolumn{1}{|c|}{\wlk \cite{wlk}} & 0.3279 & 0.7766 \\ 
				\multicolumn{1}{|c|}{\gv \cite{g2v}} & 0.4441 & 0.8272 \\ 
				\multicolumn{1}{|c|}{\svn \cite{sub2vec}} & 0.0374 & 0.1801 \\ 
				\multicolumn{1}{|c|}{\svs \cite{sub2vec}} & 0.0945 & 0.0454 \\ 
				\multicolumn{1}{|c|}{\gefsg{} \cite{gefsg}} & OOM & 0.0171 \\ \hline
			\end{tabular}%
		}
	\end{table}
	
	\subsubsection{Link prediction} \hfill\\
	\label{ss:rq1.3}
	\textbf{Dataset \& experiments.} 
	For this task, we constructed an app recommendation dataset consisting of 2,318 apps downloaded from Google Play. We build a recommendation graph $\mathcal{R}$, with these apps as nodes. An edge is placed between a pair of apps in $\mathcal{R}$, if Google Play recommends one of them while viewing the other. 
	With this graph, we follow the procedure mentioned in \cite{n2v} to cast app recommendation as a link prediction problem. 
	That is, $P$, a subset of edges (chosen at random) are removed from $\mathcal{R}$, while ensuring that this residual graph $\mathcal{R}'$ remains connected. Now, given a pair of nodes in $\mathcal{R}'$, we predict whether or not an edge exists between them. Here, endpoints of edges in $P$ are considered as positive samples and pairs of nodes with no edge between them in $R$ are considered as negative samples. We perform the experiment for $ P $ = \{10\%, 20\%, 30\%\} of total number of edges in $\mathcal{R}$. 
	
	Area under the ROC curve (AUC) is used as a metric to quantify the efficacy of link prediction.
	
	\textbf{Results \& discussions.}
	The results of the app recommendation task are presented in Table \ref{tab:rq1.3} from which the following inferences are drawn.
	
	\begin{itemize}
		[leftmargin=*]
		\setlength\itemsep{0em}
		
		\item For all values of $P$, \tool{} consistently outperforms all the baselines. Also, \tool's margin of improvement over baselines is consistent and much higher in this task than graph classification and clustering tasks. For instance, it improves best baseline performances by 3 to 4\% across all $P$ values. 
		
		\item It is noted that link prediction task does not involve any semi-supervision and hence all this improvement could be attributed to \tool's multi-view and data-driven embedding capabilities. 
		
	\end{itemize}
	
	\begin{table}[]
		\centering
		\scriptsize
		\caption{App recommendation (link prediction) results}
		\label{tab:rq1.3}
		\begin{tabular}{|c|c|c|c|}
			\hline
			\textbf{Technique} & \textbf{AUC (P = 10\%)} & \textbf{AUC (P = 20\%)} & \textbf{AUC (P = 30\%)} \\ \hline \hline
			\tool & \textbf{0.7187} & \textbf{0.7347} & \textbf{0.7236} \\ 
			\wlk \cite{wlk} & 0.6643 & 0.6865 & 0.6805 \\ 
			\gv \cite{g2v} & 0.6830 & 0.7043 & 0.6876 \\ 
			\svn \cite{sub2vec} & 0.5446 & 0.5808 & 0.5403 \\ 
			\svs \cite{sub2vec} & 0.5206 & 0.5632 & 0.5631 \\ \hline
		\end{tabular}
	\end{table}
	In sum, \tool{} consistently offers the best results across all the five tasks reported above. This illustrates that \tool's embeddings are truly task-agnostic and capture the app semantics well.

	\subsection {Single- vs. multi-view profiles}
	In this experiment, we intend to evaluate the following: (i) significance of three individual views used in \tool, (ii) significance of concatenating app profiles from individual views (i.e. linear combination), and (iii) whether non-linear combination of multiple views is better than (i) and (ii).
	
	\textbf{Dataset \& experiments.} To this end, we use the clone detection experiment reported in \S \ref{ss:rq1.2}. First, we build the app profiles (i.e., 64-dimensional embedding) with individual views (i.e., only one output layer is used in skipgram). Clone detection is then performed with each view's profile. Also, we concatenate the profiles from three views to obtain a 192-dimensional embedding and perform clone detection with the same. Finally, the regular 64-dimensional multi-view embedding from \tool{} is also used for clone detection. Due to space constraints, from this experiment onwards, only \wlk}{} is considered for comparative evaluation, as it offers the most consistent performance among the baselines considered. 

\begin{table}[t]
	\centering
	\scriptsize
	\caption{Clone detection results: single- vs. multi-view \apk{ }profiles}
	\label{tab:rq2}
	\resizebox{\columnwidth}{!}{%
		\begin{tabular}{|l|c|c|c|c|c|}
			\hline
			\multirow{2}{*}{\textbf{Technique}} & \multicolumn{5}{c|}{\textbf{Views}}                   \\ 
			& \textbf{APIs}(ARI) & \textbf{Perm.}(ARI) & \textbf{Src-sink}(ARI) & \textbf{concat.}(ARI) & \textbf{multi-view}(ARI) \\ \hline \hline
			\tool{}                  &   0.8208   &  0.7855     &  0.7953        &   0.8325      &   \textbf{0.8360}         \\ 
			\wlk \cite{wlk}       &  0.8078    &  0.7382     & 0.7479         &  0.7766       &    -        \\ \hline
		\end{tabular}%
	}
\end{table}

\textbf{Results \& discussions.} The results of this experiment are reported in Table \ref{tab:rq2}. The following inferences are drawn from the table:

\begin{itemize}
	[leftmargin=*]
	\setlength\itemsep{0em}
	\item At the outset, it is evident that individual views are capable of providing reasonable accuracies (i.e., 0.70+ ARI). This reveals that individual views possess capabilities to retain different, yet useful program semantics.
	
	\item Out of the individual views, as expected, API view yields the best accuracy.  This could be attributed to fact that this view extracts much larger number of high-quality features compared to the other two views. Owing to this well-known inference, many works in the past (e.g., \cite{drebin,casandra,droidsift,faldroid,dapasa}) have used them for a variety of tasks (incl. malware and clone detection). Also, source-sink view extract too few features to perform useful learning. In other words, it ends up underfitting the task. These observations are inline with the existing work on multi-view learning such as \cite{mkldroid}.
	
	\item In the case of \wlk, API profiles gets a very high accuracy and when they are concatenated with other views, the accuracy is reduced. We believe this is due to the inherent linearity in this mode of combination i.e., views do not complement each other.
	
	\item Interestingly, in the case of \tool, concatenating profiles from individual views yields higher accuracy than using just one view. This reveals that using multiple views is indeed offering richer semantics and helps to improve accuracy. However, concatenation could only facilitate a linear combination of views and hence is yielding slightly lesser accuracy than \tool's multi-view profiles. This illustrates the need for performing a non-linear combination of the semantic views.
	
\end{itemize}

\subsection {Semi-supervised vs. unsupervised profiling}
\label{ss:rq3}
\begin{table}[t]
	\centering
	\scriptsize
	\caption{Impact of semi-supervised embedding on malware detection efficacy}
	\label{tab:rq3}
	\setlength\tabcolsep{5pt}
	\begin{tabular}{|c|c|c|c|c|c|c|}
		\hline
		& \multicolumn{3}{c|}{\textbf{apk2vec}} & \multicolumn{3}{c|}{\textbf{\wlk \cite{wlk}}} \\ 
		\multicolumn{1}{|c|}{\textbf{Labels(\%)}} & \textbf{P(\%)}& \textbf{R(\%)} & \textbf{F(\%)} & \textbf{P(\%)} & \textbf{R(\%)} & \textbf{F(\%)} \\ \hline \hline
		\multicolumn{1}{|c|}{0} & 92.97 & 95.43 & 94.19 & 95.83 & 91.04 & 93.38 \\ 
		\multicolumn{1}{|c|}{10} & 95.44 & 97.11 & 96.27 & - & - & - \\ 
		\multicolumn{1}{|c|}{20} & 95.91 & 96.76 & 96.33 & - & - & - \\ 
		\multicolumn{1}{|c|}{30} & \textbf{96.59} & \textbf{97.28} & \textbf{96.93} & - & - & - \\ \hline
	\end{tabular}
\end{table}
In this experiment, we intend to study the impact of using the (available) class labels of apps during profiling.

\textbf{Dataset \& experiments.} 
Here, we use the same dataset which was used for online malware detection reported in \S \ref{ss:rq1.1}. 
However, in order to study the impact of varying levels of supervision, we use labels for the following percentages of samples: 10\%, 20\%, and 30\%.
Profiles for apps are built with aforementioned levels of supervision and for each setting an SVM classifier is trained and evaluated for malware detection (other settings such as train/test split are similar to \S \ref{ss:rq1.1}).

\textbf{Results \& discussions.} The results of this experiment are reported in Table \ref{tab:rq3}, from which the following inferences are made:
\begin{itemize}
	[leftmargin=*]
	\setlength\itemsep{0em}
	\item One could observe that leveraging semi-supervision during \tool's embedding is indeed helpful in improving the accuracy of the downstream task. For instance, by using labels for merely 10\% of samples help to improve the accuracy for malware detection by more than 2\%.
	
	\item Clearly, using 30\% labels yields better results than using just 10\% and 20\% labels. This illustrates the fact that the more the supervision is, the better the accuracy would be. 
	
	\item In the case of \wlk{} which uses handcrafted features, one could not use labels or other form of supervision to obtain graph embeddings. Hence, without any supervision, it performs reasonably well to obtain an f-measure of more than 93\%. However, \tool{} with even just 10\% supervision is able to outperform \wlk{} significantly i,e., by nearly 3\% f-measure. 
\end{itemize}

\subsection{Parameter Sensitivity}
The \tool{} framework involves a number of hyperparameters such as embedding dimensions ($\delta$), number of hash buckets ($B$) and number of hash functions ($k$). In this subsection, we examine how the different choices of $\delta$ affects  \tool's accuracy and efficiency, as it is the most influential hyperparameter. For the sensitivity analysis with respect to $B$ and $k$, we refer the reader to the online appendix at \cite{ourweb}. 

\textbf{Dataset \& experiments.} Here, the clone detection experiment reported in \S \ref{ss:rq1.2} is reused. The sensitivity results are fairly consistent on the remaining tasks reported in \S \ref{ss:rq1}. Except for the parameter being tested, all other parameters assume default values. Embeddings' accuracy and efficiency are determined by ARI and pretraining durations (averaged over all epochs), respectively.

\begin{figure}[t]
	\centering
	\includegraphics[height=5cm,width=8cm]{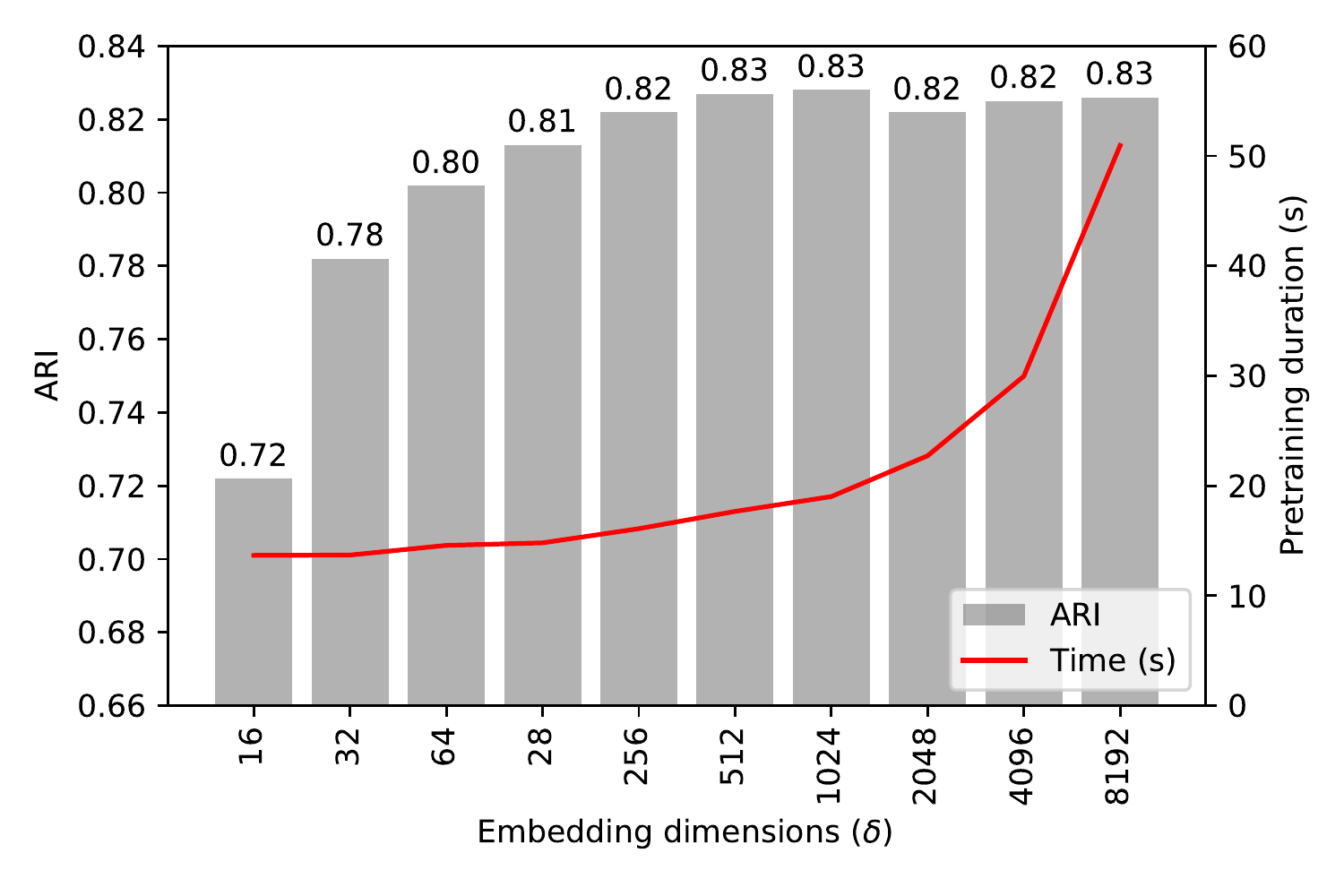}
	\caption{Sensitivity w.r.t embedding sizes 
		\label {fig:sense1}}
\end{figure}

\textbf{Results \& discussions.} These results are reported in Figure \ref{fig:sense1} from which the following inference are drawn.
\begin{itemize}
	[leftmargin=*]
	\setlength\itemsep{0em}
	\item Unsurprisingly, the ARI values increase with $\delta$. This is understandable as larger embedding sizes offer better room for learning more features. However, the performance tends to saturate once the $\delta$ is around 500 or larger. This observation is consistent with other graph \sss{} embedding approaches \cite{n2v,deepwalk,dgk}. 
	\item Also, the average pretraining time taken per epoch increases with $\delta$. This is expected, since increasing $\delta$ would result in an exponential increase in skipgram computations. This is reflected in the exponential increase in pretraining time (especially, when $\delta > 500$). This analysis helps in understanding the trade-off between \tool's accuracy and efficiency for a given dataset and picking the optimal value for $\delta$.
\end{itemize}

\section{Conclusions }
In this paper, we presented \tool, semi-supervised multimodal RL technique to automatically build data-driven behavior profiles of Android apps. Through our large-scale experiments with more than 42,000 apps, we demonstrate that profiles generated by \tool{} are task agnostic and outperform existing approaches on several tasks such as malware detection, familial clustering, clone detection and app recommendation. Our semi-supervised multimodal embeddings also prove to provide significant advantages over their unsupervised and unimodal counterparts. All the code and data used within this work is made available at \cite{ourweb}. 


\bibliographystyle{ACM-Reference-Format}

\end{document}